%% file: COT.tex
\documentclass{article}
\usepackage{float}
\usepackage{amsmath}
\usepackage{PRIMEarxiv}
\usepackage{makecell}
\usepackage[utf8]{inputenc} 
\usepackage[T1]{fontenc}    
\usepackage{hyperref}       
\usepackage{url}            
\usepackage{booktabs}       
\usepackage{amsfonts}       
\usepackage{nicefrac}       
\usepackage{microtype}      
\usepackage{lipsum}
\usepackage{fancyhdr}       
\usepackage{graphicx}       
\graphicspath{{media/}}     
\usepackage{authblk}
\usepackage{hyperref}
\usepackage{graphicx} 
\usepackage{booktabs} 
\usepackage{xcolor}   
\usepackage{pifont}   

\begin{document}
  
\title{Building a Human-Verified Clinical Reasoning Dataset via a Human–LLM Hybrid Pipeline for Trustworthy Medical AI}

\author[1†]{Chao Ding}          
\author[1†]{Mouxiao Bian}      
\author[1,2†]{Pengcheng Chen}     
\author{Hongliang Zhang}
\author[1]{Tianbin Li}
\author[1]{Lihao Liu}
\author[1]{Jiayuan Chen}
\author[3]{Zhuoran Li}
\author[3]{Yabei Zhong}
\author[3]{Yongqi Liu}
\author[4]{Haiqing Huang}
\author[4]{Dongming Shan}
\author[1]{Junjun He}
\author[1,*]{Jie Xu}
 
\affil[1]{\textit{
   Shanghai Artificial Intelligence Laboratory, \\
    Shanghai, China
}}
 
\affil[2]{\textit{
    University of Washington\\
    Washington, USA
}}
 
\affil[3]{\textit{
    Shanghai Jiao Tong University\\
     Shanghai, China
}}

\affil[4]{\textit{
    Shanghai Kupas Technology Limited Company\\
     Shanghai, China
}}
 
\footnotetext[1]{†These authors contributed equally.}
\footnotetext[2]{*Correspondence: 
Jie Xu (xujie@pjlab.org.cn)
}

\maketitle
\input{section/01_abstract}
\input{section/02_introduction}
\input{section/03_Methods}
\input{section/04_Results}
\input{section/05_Discussion}
\input{section/06_Conclusion}
\input{section/07_Additional_Information}
\bibliography{references.bib} 
\bibliographystyle{IEEEtran} 
\end{document}

%% file: section/01_abstract.tex
\begin{abstract}
Despite strong performance in medical question-answering, the clinical adoption of Large Language Models (LLMs) is critically hampered by their opaque 'black-box' reasoning, limiting clinician trust. This challenge is compounded by the predominant reliance of current medical LLMs on corpora from scientific literature or synthetic data, which often lack the granular expert validation and high clinical relevance essential for advancing their specialized medical capabilities. To address these critical gaps, we introduce a highly clinically relevant dataset with 31,247 medical question-answer pairs, each accompanied by expert-validated chain-of-thought (CoT) explanations. This resource, spanning multiple clinical domains, was curated via a scalable human-LLM hybrid pipeline: LLM-generated rationales were iteratively reviewed, scored, and refined by medical experts against a structured rubric, with substandard outputs revised through human effort or guided LLM regeneration until expert consensus. This publicly available dataset provides a vital source for the development of medical LLMs that capable of transparent and verifiable reasoning, thereby advancing safer and more interpretable AI in medicine.
\end{abstract}
\keywords{\textit{\textit{Artificial Intelligence\and Safety \and Large Language Models\and Medical AI}}}

%% file: section/02_introduction.tex
\section{Introduction}
\begin{figure}
    \centering
    \includegraphics[width=1\linewidth]{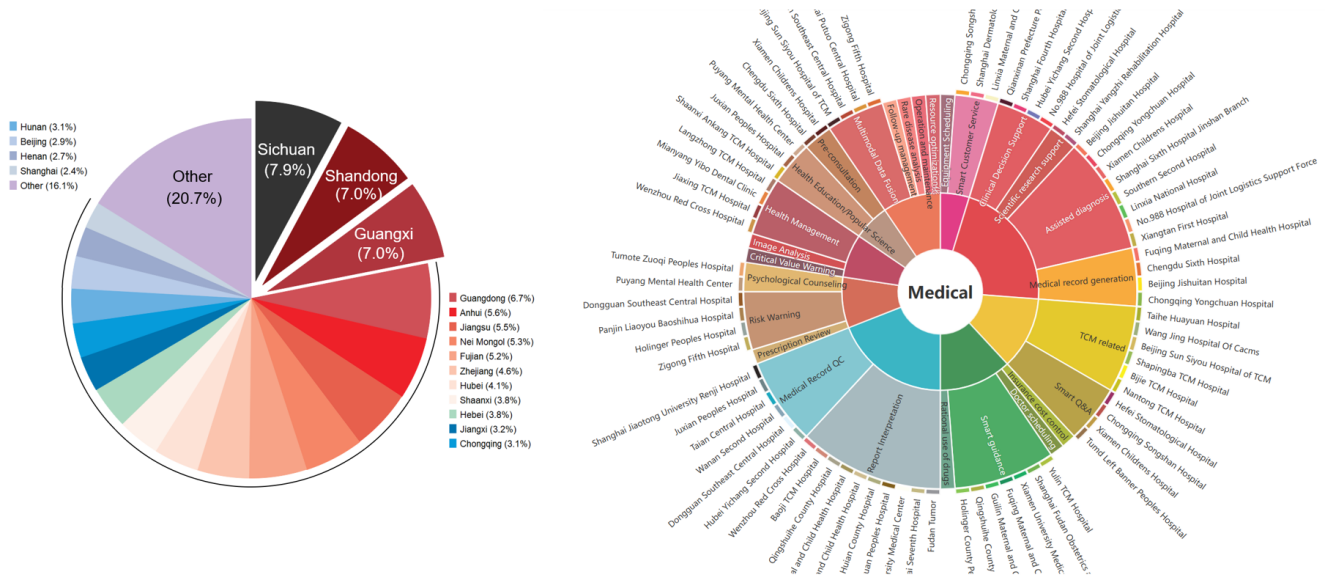}
    \caption{\textbf{Deployment Landscape of DeepSeek in the Medical Domain Across China.} This figure summarizes the adoption and application of the DeepSeek large language model in healthcare institutions nationwide. (Left): A pie chart showing the geographical distribution of over 700 medical institutions that have integrated DeepSeek. The top three provinces with the highest deployment density are Sichuan, Guangxi, and Guangdong. (Right): A radial sunburst chart illustrating the key application domains, including auxiliary diagnosis, report interpretation, intelligent triage, and medical record quality control. Each segment shows representative use cases and affiliated medical institutions across different clinical scenarios.}
    \label{Deepseek-hospital}
\end{figure}

Recent Large Language Models (LLMs) have made substantial advancements in medicine, demonstrating commendable performance in answering medical questions\cite{Touvron2023Llama}\cite{Brown2020Language}\cite{Bubeck2023Sparks}. This progress is evident in real-world applications; for instance, more than 700 Chinese medical institutions are currently using DeepSeek models, particularly in provinces such as Sichuan, Guangxi, and Guangdong. Common use cases in these settings include diagnostic assistance, report interpretation, intelligent triage, and medical record quality control \cite{Xuejun2025Digital} (Figure \ref{Deepseek-hospital}). Beyond simple knowledge recall, these models also show potential in broader clinical tasks such as documentation, triage, and patient communication\cite{Raffel2020Exploring}\cite{Chung2024Large}.

However, despite these technological capabilities and initial adoption, significant challenges hinder their large-scale integration into healthcare. Many prominent models, such as GPT-4o\cite{hurst2024gpt}, do not offer transparency into their specific reasoning processes. In the medical field, where understanding the diagnostic and treatment pathway is paramount, this 'black-box' nature makes it challenging for clinicians to confidently trust the models' judgments. Conversely, while some models, including those from the DeepSeek series (e.g., DeepSeek-R1 \cite{guo2025deepseek} and others like GPT-o1\cite{jaech2024openai}, can generate detailed chain-of-thought explanations, the content of these CoTs is often primarily derived from reinforcement learning and typically lacks rigorous expert review. This casts considerable doubt on their reliability and clinical trustworthiness. Consequently, the large-scale adoption of AI in healthcare remains limited, primarily due to these fundamental issues of trust and the verifiability of their reasoning processes\cite{Brin2023Comparing}\cite{Yaneva2024Examining} \cite{Thirunavukarasu2023Large}.

To address the challenges associated with incorporating robust reasoning into medical training data, several approaches have been investigated. For instance, the Huatuo-o1 CoT\cite{chen2024huatuogpt} leveraged GPT-4o to transform verifiable medical questions from the MedQA dataset into formats suitable for CoT and reinforcement learning. However, due to the inherent performance limitations of GPT-4o itself, the accuracy of this transformation process could not be entirely assured. Another notable effort, MedReason\cite{wu2025medreason}, implemented a sophisticated pipeline encompassing entity alignment, knowledge graph reasoning path identification, LLM-driven text expansion, and answer verification. This pipeline was designed to automatically convert raw clinical question-answering (QA) pairs into structured, verifiable medical Chains-of-Thought. However, a limitation of the MedReason methodology was that clinical experts were predominantly engaged only during the final manual verification phase. In this stage, their responsibilities included sample-based quality control and subjective assessments. Physicians were not directly involved in the earlier critical stages of the workflow, such as automatic data generation, filtering, or the model training process itself.

In this work, we introduce a novel human-AI collaborative workflow designed to ensure clinical experts are deeply involved throughout the entire dataset construction process, rather than being limited to a role of mere post-hoc evaluation.The multi-stage process of this pipeline, from initial data sourcing and LLM-based expansion to iterative expert review, AI re-verification, and consensus-driven validation, is illustrated in Figure \ref{full-pipeline}. Leveraging this expert-centric framework, we have generated a dataset of 30,000 medical QA pairs. This comprehensive dataset covers the vast majority of medical domains, and our methodology emphasizes producing high-quality, clinically relevant data at scale, building upon principles of expert-driven curation seen in resources like EXPERTQA \cite{Malaviya2024ExpertQA} while distinctively focusing on the deep, iterative co-creation with medical specialists. The comparison of our dataset and the other work is demonstrated in Table \ref{tab:data_comparison}

Our contributions include:
\begin{itemize}
\item Comprehensive Dataset: A medical QA dataset to integrate large-scale CoT explanations with multi-tier experts deeply involved in data generation and validation, ensuring clinical accuracy and relevance.
\item Dynamic Curation Pipeline: A human-AI workflow combining LLM-generated drafts, automated question augmentation, multi-pass expert review, and consensus-driven corrections. This approach leverages AI scalability and human expertise to produce trusted content.
\item  Multi-Dimensional Evaluation: A structured rubric assessing explanations across five dimensions—medical correctness, reasoning structure, information sufficiency, terminology clarity, and clinical value—aligned with best practices for safe AI evaluation. These annotations are included in the dataset, enabling robust training and benchmarking.
\item Empirical Validation: Baseline experiments demonstrate that models fine-tuned on our expert-validated dataset outperform those trained on unvalidated data, achieving higher answer accuracy and expert-rated reasoning quality. These findings align with prior studies showing the benefits of expert-curated data for QA performance
\end{itemize}
In the following sections, we describe the dataset construction, human-AI workflow, and content quality analysis. We also present experimental results showcasing the dataset’s impact on medical QA systems. Our approach underscores the importance of rigorous dataset curation in bridging the reliability gap between AI and human experts in healthcare, paving the way for transparent and trustworthy medical AI.

\vspace{-3mm}
\begin{table}[t]
\setlength{\tabcolsep}{1.2mm} 
  \centering
\scalebox{0.80}{
    \begin{tabular}{lccccc} 
    \toprule
    \textbf{Data Source} &  \textbf{Quality Filtering} & \textbf{Medical Specific} & \textbf{Factual Guidance} & \textbf{Expert Checking} & \textbf{Expert Involved Data Generation} \\ 
    \midrule
    Distillation & \textcolor{red}{\ding{55}}    & \textcolor{red}{\ding{55}}    & \textcolor{red}{\ding{55}}    & \textcolor{red}{\ding{55}} & \textcolor{red}{\ding{55}} \\ 
    BOLT  & \textcolor{green}{\ding{51}}   & \textcolor{red}{\ding{55}}    & \textcolor{red}{\ding{55}}    & \textcolor{red}{\ding{55}} & \textcolor{red}{\ding{55}} \\ 
    Huatuo-o1 CoT & \textcolor{green}{\ding{51}}   & \textcolor{green}{\ding{51}}   & \textcolor{red}{\ding{55}}    & \textcolor{red}{\ding{55}} & \textcolor{red}{\ding{55}} \\ 
    MedReason & \textcolor{green}{\ding{51}}   & \textcolor{green}{\ding{51}}   & \textcolor{green}{\ding{51}}   & \textcolor{green}{\ding{51}} & \textcolor{red}{\ding{55}} \\ 
    \textbf{Ours*} & \textcolor{green}{\ding{51}}   & \textcolor{green}{\ding{51}}   & \textcolor{green}{\ding{51}}   & \textcolor{green}{\ding{51}} & \textcolor{green}{\ding{51}} \\ 
    \bottomrule
    \end{tabular}%
    }
    \caption{\textbf{Comparison of Chain-of-Thought (CoT) data sources.} As demonstrated in the table, our approach comprehensively incorporates all evaluated quality features: quality filtering, medical specificity, factual guidance, expert checking, and direct expert involvement in the data generation process. This holistic methodology yields high-quality medical CoT, with its overall quality further affirmed through assessment by medical experts.} 
  \label{tab:data_comparison}%
  \vspace{-6mm}
\end{table}%

%% file: section/03_Methods.tex
\section{Methods}
\subsection{Source Data and Initial Question Set}
We used 3,621 questions from the medical examination dataset https://medbench.opencompass.org.cn/docs. The questions were objective multiple-choice and were designed as K-type (best-choice) and S-type (best-choice for medical record summarization) questions, with brief descriptions of the following two types of questions based on multiple-choice questions. These questions cover all major medical disciplines: internal medicine, surgery, pediatrics, obstetrics and gynecology, preventive medicine, etc., reflecting the breadth of competencies required of general practitioners. Each question is a single-choice question (one correct answer out of four or five options) covering a wide range of topics from basic biomedical knowledge to clinical diagnosis and treatment. To ensure diversity and representativeness, our data were derived from real clinical diagnosis and treatment scenarios so that the distribution of test questions corresponded to actual clinical departments. Our dataset covers a wide range of medical knowledge domains with sufficient depth in each domain. (supplementary material)
\subsection{LLM-Generated Reasoning and Dataset Expansion}
We used a large language model (DeepSeek-R1) for each seed question to generate a detailed CoT. The model is prompted to “think out loud”, explain why each answer choice is correct or incorrect, and ultimately choose the correct answer. This step generates rich inference trajectories for each question, effectively transforming simple Q\&A pairs into more informative triples - question, answer, and explanation. Next, we leveraged these model-generated explanations to synthesize new QA pairs. The goal was to create novel questions that remain rooted in the original medical concepts but present them in different ways, thereby expanding the dataset tenfold without requiring entirely new manual questions. We used an automated question generation approach wherein the model’s explanation for a given question was parsed to identify key facts and reasoning steps. These were then reformulated into new question prompts. For instance, if an original question asked about the primary indication for a drug (with the explanation covering why the drug is suitable for certain patients but not others), we could generate a new question focusing on a different aspect of that explanation (e.g., contraindications or mechanism of action). We also permuted clinical vignettes and numerical values where applicable, to add variety while preserving clinical validity. Each synthesized question was answered by the model (providing both an initial answer and a reasoning path). This procedure expanded the dataset from 3,621 to ~36,210 QA pairs. Importantly, at this stage, the data was still unverified – the model’s outputs likely included some inaccuracies or unnatural phrasing. Thus, we proceeded to an extensive validation and refinement phase to ensure quality and trustworthiness.
\subsection{Multi-Stage Expert–AI Validation Pipeline}
\begin{figure}
    \centering
    \includegraphics[width=1\linewidth]{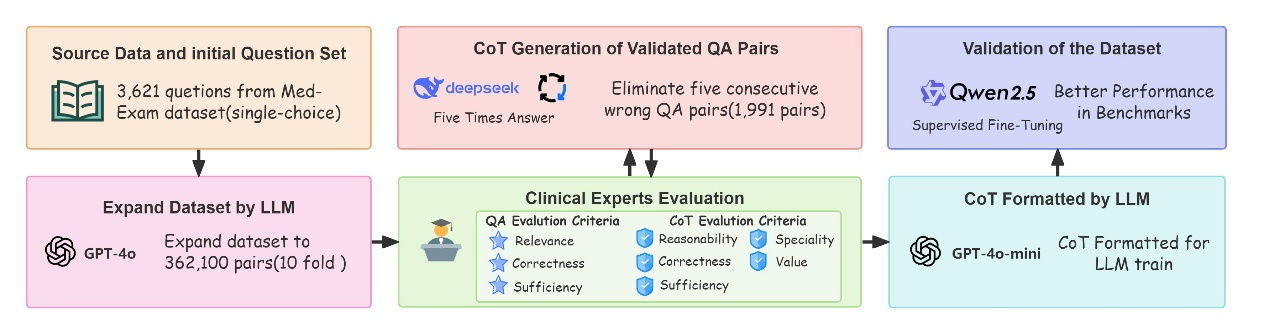}
    \caption{\textbf{Human–LLM hybrid pipeline for constructing a clinically validated medical QA dataset with reasoning chains.} The figure outlines the multi-stage workflow used to generate and validate a large-scale Chinese medical QA dataset. Starting from 3,621 seed questions drawn from national medical examinations, DeepSeek-R1 was prompted to generate step-by-step chain-of-thought (CoT) explanations. These rationales were used to synthesize 30,000 QA pairs. Medical experts then reviewed the QA items for correctness and clarity. Each item was re-answered by the LLM up to five times to detect formulation flaws. Items triggering the five-strike error mechanism underwent expert panel review, where CoTs were re-generated and scored across five dimensions: medical correctness, reasoning structure, information sufficiency, terminology clarity, and clinical utility. The validated dataset serves as a benchmark for training and evaluating trustworthy medical AI systems.}
    \label{full-pipeline}
\end{figure}
\subsubsection{Initial Human Review}
To ensure the clinical validity, clarity, and educational value of the QA pairs, a structured human validation phase was conducted by a team comprising practicing physicians and advanced medical trainees (see supplementary material). Each of the 36,210 question–answer pairs was independently reviewed according to a standardized evaluation protocol. Reviewers assessed three primary dimensions: whether the answer directly addressed the question (relevance), whether the answer was factually accurate and aligned with contemporary medical guidelines (medical correctness), and whether the answer provided sufficiently complete and clinically meaningful information (information sufficiency). Scoring was binary, with each pair assigned a score of 1 (correct and sufficient) or 0 (incorrect, incomplete, or irrelevant).
\subsubsection{AI Re-Answering and Verification}
After a human reviewer cleaned up the questions and answers, we answered each question again using DeepSeek-R1 (without providing edited explanations). We recorded whether the model answered correctly and checked its new inference chain. If DeepSeek-R1 answered correctly and reasonably, it indicates that the question is likely to be formulated. If it answered incorrectly, it indicated a potential problem - perhaps the question was still ambiguous, too tricky, or beyond the model's knowledge. Rather than dropping such questions outright, we used these failures as indicators for further review.
\subsubsection{Five-Strike Error Trigger for Expert Panel}
We allowed the model up to five attempts to get each question right. This “five-strike” rule is a stringent filter: if the AI failed to produce the correct answer after five tries, the item was deemed challenging or flawed and was escalated to a panel of medical experts for in-depth review. The expert panel consisted of experienced clinicians (e.g., attending physicians or subject-matter experts in various specialties). Approximately 6.4\% of the questions triggered the five-strike mechanism. These tended to be questions with subtle distinctions or those requiring specialized knowledge not well covered in general sources.
\subsubsection{Expert Panel Refinement}
To further ensure clinical validity and reasoning quality, we implemented a formal expert review phase for problematic QA-CoT pairs identified during automated validation. Items that triggered the five-strike error criterion—defined as five consecutive failures by the model to select the correct answer under varied prompting—were escalated to an expert panel for detailed adjudication.
Each flagged item, along with its model answer history and intermediate human reviews, was independently assessed by at least two licensed medical experts following the MedCoT Quality Annotation Standard v1.0 protocol (see supplementary material). Experts evaluated each item across five critical dimensions: medical correctness, reasoning structure, information sufficiency, terminology clarity, and clinical utility, using a 0–2 scoring rubric for each dimension.
\subsubsection{Clinical Consensus Approval}
Finally, every question – whether it went through the extra expert panel step or not – was approved by consensus. This meant at least two experts agreed on the correct answer and phrasing, and that the question aligned with current clinical knowledge/practice. Items that could not reach consensus were removed or replaced. This step guarantees that every QA pair in the dataset is clinically accurate and relevant.

%% file: section/04_Results.tex
\section{Results}
\subsection{Dataset Content and Statistics}
\begin{figure}
    \centering
    \includegraphics[width=1\linewidth]{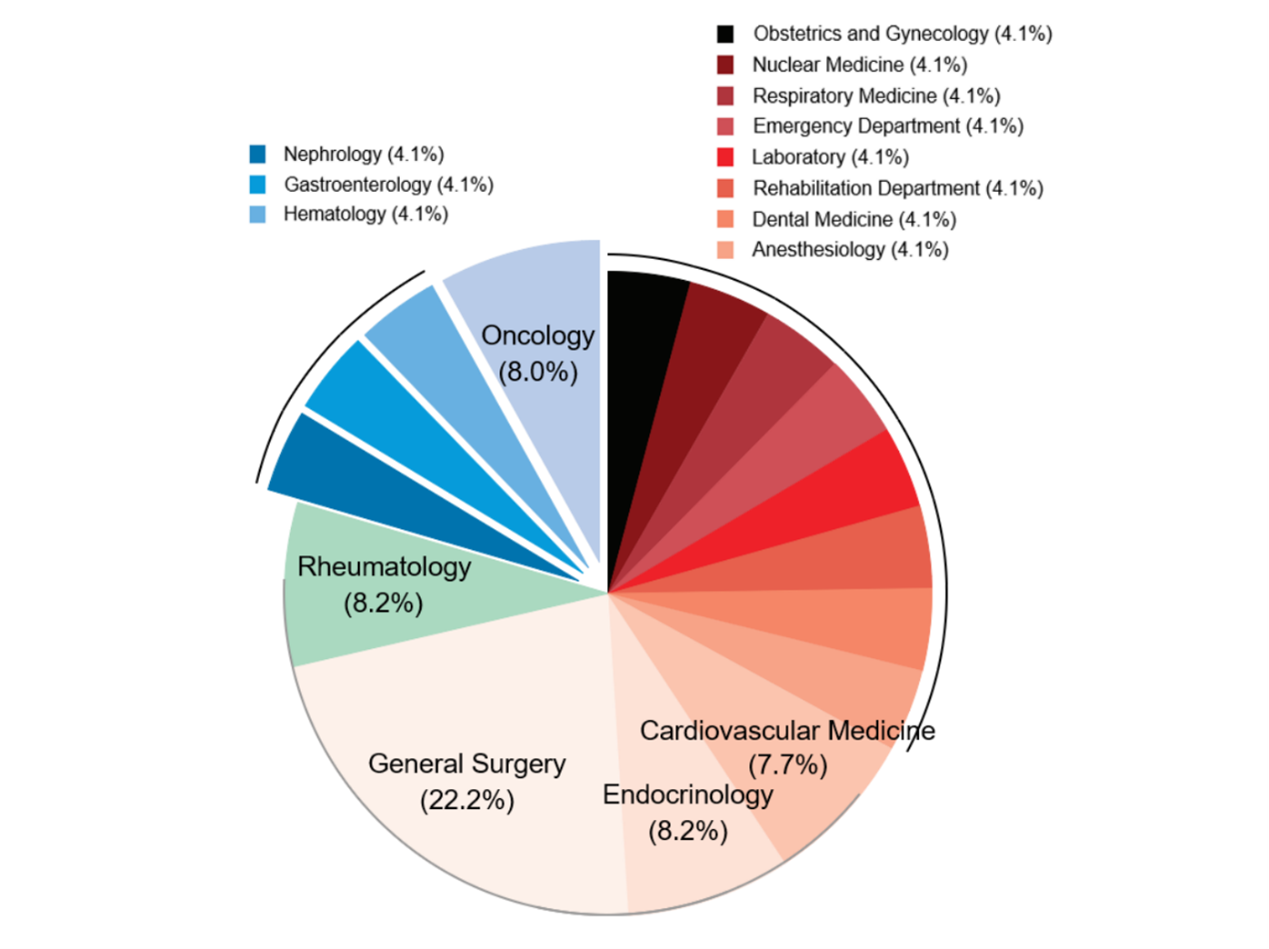}
    \caption{\textbf{Distribution of clinical specialties represented in the medical QA dataset.} The pie chart shows the proportional representation of 12 medical disciplines in the dataset. General Surgery accounts for 22.2\% of all items, while Rheumatology, Endocrinology, and Oncology each contribute between 8–8.2\%. A range of other core departments—including Cardiovascular Medicine, Nephrology, Gastroenterology, Hematology, and Emergency Services—are also represented, ensuring clinical breadth across both internal medicine and procedural specialties.}
    \label{piechart-distribution of clinical specialites}
\end{figure}

Our final dataset consists of 30,000 single-choice medical QA items, each comprising: a question (in Chinese), 4–5 answer options, the correct answer, and a chain-of-thought explanation. All core fields of medicine are well-represented. For example, internal medicine contributes the largest share with X\% of questions (covering cardiology, pulmonology, gastroenterology, etc.), surgery 22.2\%, endocrinology 8.2\%, rheumatology 8.2\%, oncology 8.0\%, oncology 8.0\%, and so on, collectively spanning dozens of subtopics (Figure \ref{piechart-distribution of clinical specialites}). The diversity of questions is illustrated by the range of formats: some are straightforward knowledge recall (e.g., diagnostic criteria, drug mechanisms), while others are clinical vignettes requiring application of knowledge to a patient scenario. We also categorized questions by cognitive level (recall, interpretation, problem-solving) and difficulty (easy, moderate, hard) as judged by our experts during validation.

\subsection{Quality of Explanations (Rubric Evaluation)}
\begin{figure}
    \centering
    \includegraphics[width=0.5\linewidth]{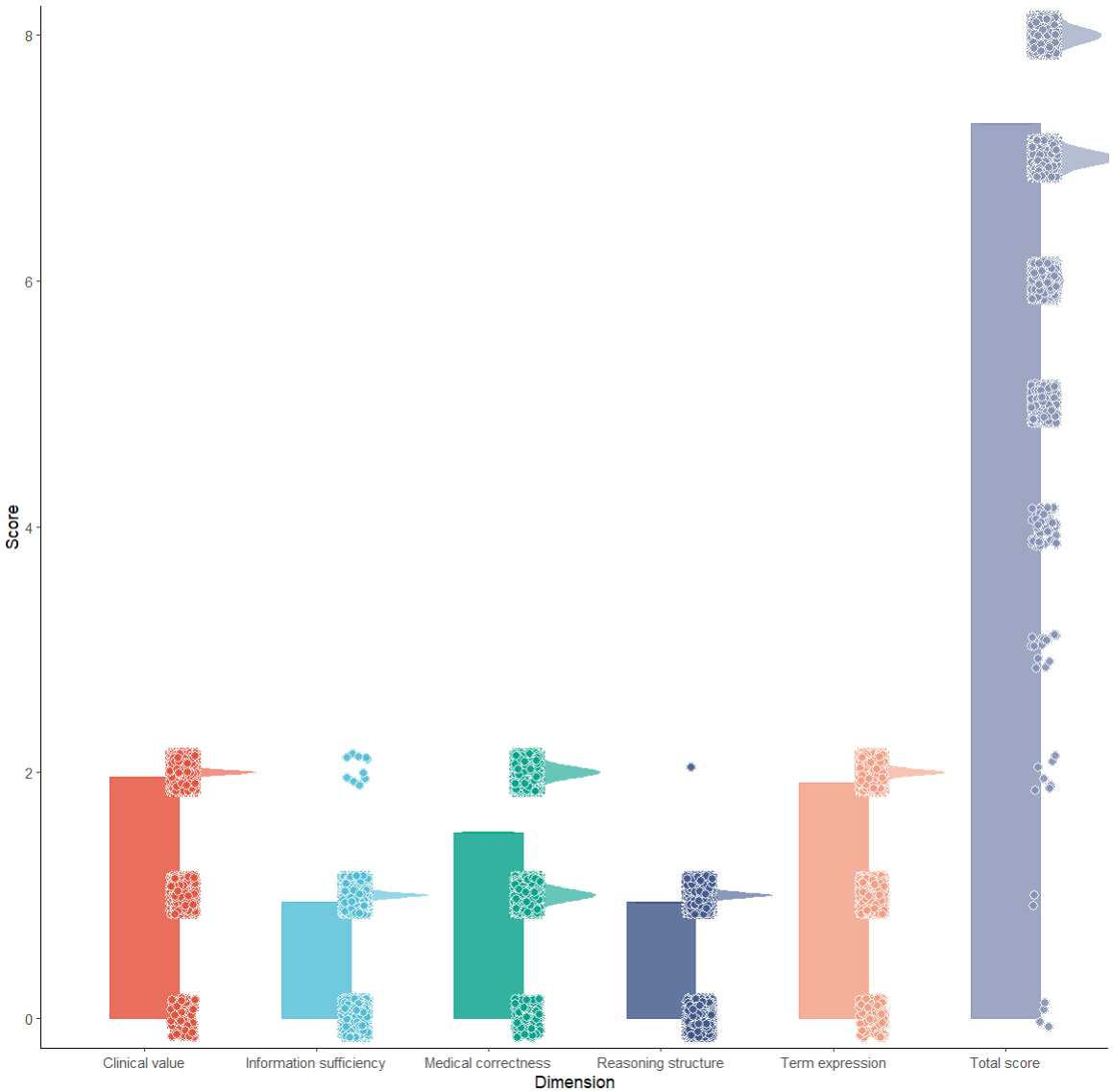}
    \caption{\textbf{Five-dimensional expert evaluation of CoT quality.}Visualizes the five-dimensional quality assessment of CoT explanations in the constructed medical QA dataset. Each CoT instance was scored across five expert-defined dimensions: medical correctness, reasoning structure, information sufficiency, terminology clarity, and clinical utility.}
    \label{Five-dimensional expert evaluation of CoT quality}
\end{figure}

We constructed a total of 36,213 QA pairs with corresponding chain-of-thought (CoT) explanations. During expert validation, 34,062 QA pairs were confirmed to be medically correct and clinically valuable. To assess the robustness of each item, we applied a five-strike re-answering mechanism using DeepSeek-R1; 6.4\% of the questions (n = 1,991) failed all five model attempts and were flagged for expert panel review. A total of 32,071 formatted CoT rationales were finalized. Among them, 25,989 (81\%) required no revision after expert inspection, while 5,258 (16\%) were modified and 824 (3\%) were discarded due to irreparable issues. All questions were reviewed by at least two medical experts, and 100\% consensus was reached before inclusion. Final CoT quality was assessed using a five-dimensional rubric (Figure \ref{Five-dimensional expert evaluation of CoT quality}).
\subsection{Baseline Model Performance and Utility of the Dataset}

%% file: section/05_Discussion.tex
\section{Discussion}
While AI and LLM are often designed as aids to support clinical decision making (e.g., diagnostic AI, clinical monitoring, etc.)\cite{Tu2024Towards}\cite{Chen2025Framework}\cite{Barnett2023Real},unreliable, unsafe, and unexplained black-box AI models can lead to misdiagnosis of disease and subsequent mistreatment. In this work, we present a high-quality healthcare quality assurance dataset built through a collaborative AI pipeline. Our results show that this approach is capable of providing inspiring ideas for building trustworthy healthcare AI systems. Ensuring the reliability of medical datasets is fundamental to the safe deployment of AI, similar to the role of clinical trials in therapy. Traditional data collection methods, such as scouring data from a library of test papers or using crowdsourced annotators, can hide the risk of inaccuracies. By involving human experts in multiple rounds of review, we ensured that the dataset can be trusted as a source of truth\cite{Alber2025Medical}\cite{Wang2024Safety}. Ensuring the reliability of datasets is a foundational step toward building trustworthy AI systems, much like rigorous clinical trials are necessary for medical therapeutics\cite{Schwabe2024METRIC}.Even state-of-the-art(SOTA) proprietary LLMs perpetuate historical bias\cite{Omiye2023Large}.
Dynamic Human–AI Workflow Benefits: Our findings highlight the synergy between AI and expert humans. The AI (DeepSeek-R1) greatly accelerated data generation, producing tens of thousands of plausible QA items that would have taken experts alone an immense effort to create from scratch. At the same time, human expertise was crucial in refining this AI-generated content – not only to fix errors, but also to impart clinical realism and relevance that a generative model might not fully grasp. The iterative loop (AI proposes, humans dispose or adjust, AI checks again) is reminiscent of a knowledge refinement cycle, where each pass makes the data more robust. Interestingly, using the AI as a tool to test the questions (the re-answering stage) provided an innovative form of validation – essentially a “AI adversarial test”. If a state-of-the-art model can’t reliably answer a question, that either signals a very hard question or a flaw; in both cases, scrutinizing such questions led to improvements. This dynamic workflow can be seen as a blueprint for other domains requiring high assurance data. It shows that AI and humans, working together, can produce better outcomes than either alone – AI contributes breadth and speed, while humans contribute judgment and precision\cite{Malaviya2024ExpertQA}.
We placed a strong emphasis not just on what the model answered, but how it reasoned. This aligns with a broader movement in AI toward evaluating models on criteria like explainability, fairness, and safety. By designing a rubric and scoring the explanations, we obtained insights into the strengths and weaknesses of AI-generated reasoning. For example, our analysis revealed that information sufficiency was an area for potential improvement – a point that might have been overlooked if we only checked final answers. The rubric approach also provides a form of structured feedback that could be used to further tune models (e.g., via reinforcement learning or by developing automated evaluators that predict these scores). Moreover, this kind of evaluation is directly useful to end-users: clinicians are more likely to trust an AI whose answers come with convincing, transparent rationales. We have effectively built a dataset that can help train models to produce those rationales and a methodology to judge them, contributing to the explainability aspect of trustworthy AI.
Despite our best efforts, there are limitations to acknowledge. First, the dataset is in Chinese and focused on the context of Chinese medical scenario. While many medical principles are universal, some guidelines or practices might be country-specific. Users aiming to apply this dataset in other contexts. However, our approach is language-independent and can be extended to other languages and bilingual experts. Second, the initial generation phase relied solely on DeepSeek-R1, which may introduce model-specific biases in content style or topic coverage. While expert review helped mitigate overt errors, subtle artifact may persist. Employing multiple generative models or ensembles could help diversify future dataset development. Third, expert validation is resource-intensive. The iterative review process requires substantial human effort, particularly when assembling panels of qualified clinicians. While feasible at the current scale, expanding the dataset significantly may pose logistical challenges. Future workflows may benefit from automated assessments to reduce this burden. Finally, although the dataset is large, it is not comprehensive. Medicine is an ever-evolving field, and in order to remain clinically relevant, we plan to keep updating this research program.

%% file: section/06_Conclusion.tex
\section{Conclusion}
We present the first large-scale, Chinese-language medical QA dataset with expert-validated reasoning chains, developed through a rigorous human–AI collaborative pipeline. By combining LLM generation with multi-stage expert review, the dataset establishes a new standard for trustworthy training and evaluation in medical AI. It supports the development of models with more accurate, interpretable, and clinically aligned reasoning capabilities. We release this resource to catalyze further research in safe and explainable medical AI, and invite the community to build upon it.

%% file: section/07_Additional_Information.tex
\section{Additional Information}
\textbf{\textbf{Data availability}}

All the data generated in this study is available at the online repository: 

\href{https://medbench.opencompass.org.cn/community/data-station}{https://medbench.opencompass.org.cn/community/data-station} 

\textbf{\textbf{Code availability}}

 \textbf{\textbf{Author contributions}}

C.D., M.X.B., H.L.Z, T.B.L., C.P.C , L.H.L., and J.X.conceived the study. H.L.Z, T.B.L., C.P.C and J.X. designed the study, collected data, and conducted data analyses. Z.Y.B., Y.Q.L., and J.Y.C. drafted the manuscript. K
Z.R.L., C.P.C, L.H.L., and J.X.supervised the study. All authors have read and approved the manuscript.

\textbf{\textbf{Competing interests}}

The authors declare no competing interests.

\textbf{\textbf{Additional information}}

Correspondence and requests for materials should be addressed to Jie Xu.

\textbf{\textbf{Acknowledgements}}

Supported by Shanghai Artificial Intelligence Laboratory

%% file: COT.bbl
\begin{thebibliography}{10}
\providecommand{\url}[1]{#1}
\csname url@samestyle\endcsname
\providecommand{\newblock}{\relax}
\providecommand{\bibinfo}[2]{#2}
\providecommand{\BIBentrySTDinterwordspacing}{\spaceskip=0pt\relax}
\providecommand{\BIBentryALTinterwordstretchfactor}{4}
\providecommand{\BIBentryALTinterwordspacing}{\spaceskip=\fontdimen2\font plus
\BIBentryALTinterwordstretchfactor\fontdimen3\font minus \fontdimen4\font\relax}
\providecommand{\BIBforeignlanguage}[2]{{%
\expandafter\ifx\csname l@#1\endcsname\relax
\typeout{** WARNING: IEEEtran.bst: No hyphenation pattern has been}%
\typeout{** loaded for the language `#1'. Using the pattern for}%
\typeout{** the default language instead.}%
\else
\language=\csname l@#1\endcsname
\fi
#2}}
\providecommand{\BIBdecl}{\relax}
\BIBdecl

\bibitem{Touvron2023Llama}
H.~Touvron, T.~Lavril, G.~Izacard \emph{et~al.}, ``Llama: Open and efficient foundation language models,'' \emph{arXiv preprint arXiv:2302.13971}, 2023.

\bibitem{Brown2020Language}
T.~Brown, B.~Mann, N.~Ryder \emph{et~al.}, ``Language models are few-shot learners,'' \emph{Advances in neural information processing systems}, vol.~33, pp. 1877--1901, 2020.

\bibitem{Bubeck2023Sparks}
S.~Bubeck and et~al., ``Sparks of artificial general intelligence: Early experiments with gpt-4,'' ArXiv, 2023.

\bibitem{Xuejun2025Digital}
Y.~Xue~jun, J.~I.~A. Wang, M.~A.~O. Ying \emph{et~al.}, ``Digital and artificial intelligence technologies empowering neurosurgery,'' \emph{Chinese Journal of Contemporary Neurology \& Neurosurgery}, vol.~25, no.~2, 2025.

\bibitem{Raffel2020Exploring}
C.~Raffel, N.~Shazeer, A.~Roberts \emph{et~al.}, ``Exploring the limits of transfer learning with a unified text - to - text transformer,'' \emph{Journal of machine learning research}, vol.~21, no. 140, pp. 1--67, 2020.

\bibitem{Chung2024Large}
P.~Chung and et~al., ``Large language model capabilities in perioperative risk prediction and prognostication,'' \emph{JAMA surgery}, vol. 159, pp. 928--937, 2024.

\bibitem{hurst2024gpt}
A.~Hurst, A.~Lerer, A.~P. Goucher, A.~Perelman, A.~Ramesh, A.~Clark, A.~Ostrow, A.~Welihinda, A.~Hayes, A.~Radford \emph{et~al.}, ``Gpt-4o system card,'' \emph{arXiv preprint arXiv:2410.21276}, 2024.

\bibitem{guo2025deepseek}
D.~Guo, D.~Yang, H.~Zhang, J.~Song, R.~Zhang, R.~Xu, Q.~Zhu, S.~Ma, P.~Wang, X.~Bi \emph{et~al.}, ``Deepseek-r1: Incentivizing reasoning capability in llms via reinforcement learning,'' \emph{arXiv preprint arXiv:2501.12948}, 2025.

\bibitem{jaech2024openai}
A.~Jaech, A.~Kalai, A.~Lerer, A.~Richardson, A.~El-Kishky, A.~Low, A.~Helyar, A.~Madry, A.~Beutel, A.~Carney \emph{et~al.}, ``Openai o1 system card,'' \emph{arXiv preprint arXiv:2412.16720}, 2024.

\bibitem{Brin2023Comparing}
D.~Brin and et~al., ``Comparing chatgpt and gpt-4 performance in usmle soft skill assessments,'' \emph{Scientific reports}, vol.~13, p. 16492, 2023.

\bibitem{Yaneva2024Examining}
V.~Yaneva, P.~Baldwin, D.~Jurich, K.~Swygert, and B.~Clauser, ``Examining chatgpt performance on usmle sample items and implications for assessment,'' \emph{Academic medicine : journal of the Association of American Medical Colleges}, vol.~99, pp. 192--197, 2024.

\bibitem{Thirunavukarasu2023Large}
A.~Thirunavukarasu and et~al., ``Large language models in medicine,'' \emph{Nature medicine}, vol.~29, pp. 1930--1940, 2023.

\bibitem{chen2024huatuogpt}
J.~Chen, Z.~Cai, K.~Ji, X.~Wang, W.~Liu, R.~Wang, J.~Hou, and B.~Wang, ``Huatuogpt-o1, towards medical complex reasoning with llms,'' \emph{arXiv preprint arXiv:2412.18925}, 2024.

\bibitem{wu2025medreason}
J.~Wu, W.~Deng, X.~Li, S.~Liu, T.~Mi, Y.~Peng, Z.~Xu, Y.~Liu, H.~Cho, C.-I. Choi \emph{et~al.}, ``Medreason: Eliciting factual medical reasoning steps in llms via knowledge graphs,'' \emph{arXiv preprint arXiv:2504.00993}, 2025.

\bibitem{Malaviya2024ExpertQA}
C.~Malaviya and et~al., ``Expertqa: Expert-curated questions and attributed answers,'' in \emph{Association for Computational Linguistics}, Mexico City, Mexico, 2024, pp. 3025--3045.

\bibitem{Tu2024Towards}
T.~Tu and et~al., ``Towards conversational diagnostic ai,'' 2024.

\bibitem{Chen2025Framework}
E.~Chen, S.~Prakash, V.~Janapa~Reddi, D.~Kim, and P.~Rajpurkar, ``A framework for integrating artificial intelligence for clinical care with continuous therapeutic monitoring,'' \emph{Nature Biomedical Engineering}, vol.~9, pp. 445--454, 2025.

\bibitem{Barnett2023Real}
M.~Barnett and et~al., ``A real-world clinical validation for ai-based mri monitoring in multiple sclerosis,'' \emph{NPJ digital medicine}, vol.~6, p. 196, 2023.

\bibitem{Alber2025Medical}
D.~Alber and et~al., ``Medical large language models are vulnerable to data-poisoning attacks,'' \emph{Nature medicine}, vol.~31, pp. 618--626, 2025.

\bibitem{Wang2024Safety}
X.~Wang and et~al., ``Safety challenges of ai in medicine,'' 2024.

\bibitem{Schwabe2024METRIC}
D.~Schwabe, K.~Becker, M.~Seyferth, A.~Klaß, and T.~Schaeffter, ``The metric-framework for assessing data quality for trustworthy ai in medicine: a systematic review,'' \emph{NPJ digital medicine}, vol.~7, p. 203, 2024.

\bibitem{Omiye2023Large}
J.~Omiye, J.~Lester, S.~Spichak, V.~Rotemberg, and R.~Daneshjou, ``Large language models propagate race-based medicine,'' \emph{NPJ digital medicine}, vol.~6, p. 195, 2023.

\end{thebibliography}
